\documentclass[letterpaper, 10 pt, conference]{ieeeconf}  

\IEEEoverridecommandlockouts                              

\usepackage{amsmath,amsthm,amssymb}
\usepackage{mathtools}
\mathtoolsset{showonlyrefs}
\usepackage{mathrsfs}
\usepackage{comment}
\usepackage{multicol,multirow}
\usepackage[dvipsnames,table]{xcolor}
\usepackage{caption}
\usepackage{subcaption}
\usepackage{graphicx}
\usepackage{multirow}
\usepackage{cite}
\usepackage[inkscapeformat=png]{svg}

\usepackage[ruled,linesnumbered]{algorithm2e}

    \makeatletter
    \let\NAT@parse\undefined
    \makeatother

\usepackage[hidelinks]{hyperref}

\LinesNumbered

\title{\LARGE \bf
Online Update of Safety Assurances Using Confidence-Based Predictions
}
\author{Kensuke Nakamura$^1$ and Somil Bansal$^2$
\thanks{\noindent $^1$Author is with the MAE Department at Princeton University; \href{mailto:k.nakamura@princeton.edu}{\tt\footnotesize k.nakamura@princeton.edu}.
$^2$Author is with the ECE department at USC; \href{mailto:somilban@usc.edu}{\tt\footnotesize somilban@usc.edu.} Project website: \url{https://kensukenk.github.io/OnlineConfidenceUpdate/}\newline This research is supported in part by the NVIDIA Academic Hardware Grant Program and the USC SURE Program.}
}

\renewcommand{\baselinestretch}{0.987}

\begin{document}

\maketitle
\thispagestyle{empty}
\pagestyle{empty}

\begin{abstract}
Robots such as autonomous vehicles and assistive manipulators are increasingly operating in dynamic environments and close physical proximity to people. 
In such scenarios, the robot can leverage a human motion predictor to predict their future states and plan safe and efficient trajectories.
However, no model is ever perfect -- when the observed human behavior deviates from the model predictions, the robot might plan unsafe maneuvers.
Recent works have explored maintaining a confidence parameter in the human model to overcome this challenge, wherein the predicted human actions are tempered online based on the likelihood of the observed human action under the prediction model.
This has opened up a new research challenge, i.e., \textit{how to compute the future human states online as the confidence parameter changes?}
In this work, we propose a Hamilton-Jacobi (HJ) reachability-based approach to overcome this challenge.
Treating the confidence parameter as a virtual state in the system, we compute a parameter-conditioned forward reachable tube (FRT) that provides the future human states as a function of the confidence parameter. 
Online, as the confidence parameter changes, we can simply query the corresponding FRT, and use it to update the robot plan.
Computing parameter-conditioned FRT corresponds to an (offline) high-dimensional reachability problem, which we solve by leveraging recent advances in data-driven reachability analysis.
Overall, our framework enables online maintenance and updates of safety assurances in human-robot interaction scenarios, even when the human prediction model is incorrect.
We demonstrate our approach in several safety-critical autonomous driving scenarios, involving a state-of-the-art deep learning-based prediction model. 
\end{abstract}

\section{Introduction} \label{sec:intro}

\begin{figure}[t]
\centering
\vspace{-2em}
\includegraphics[trim={5cm 0 5cm 0},clip, width=0.98\columnwidth]
{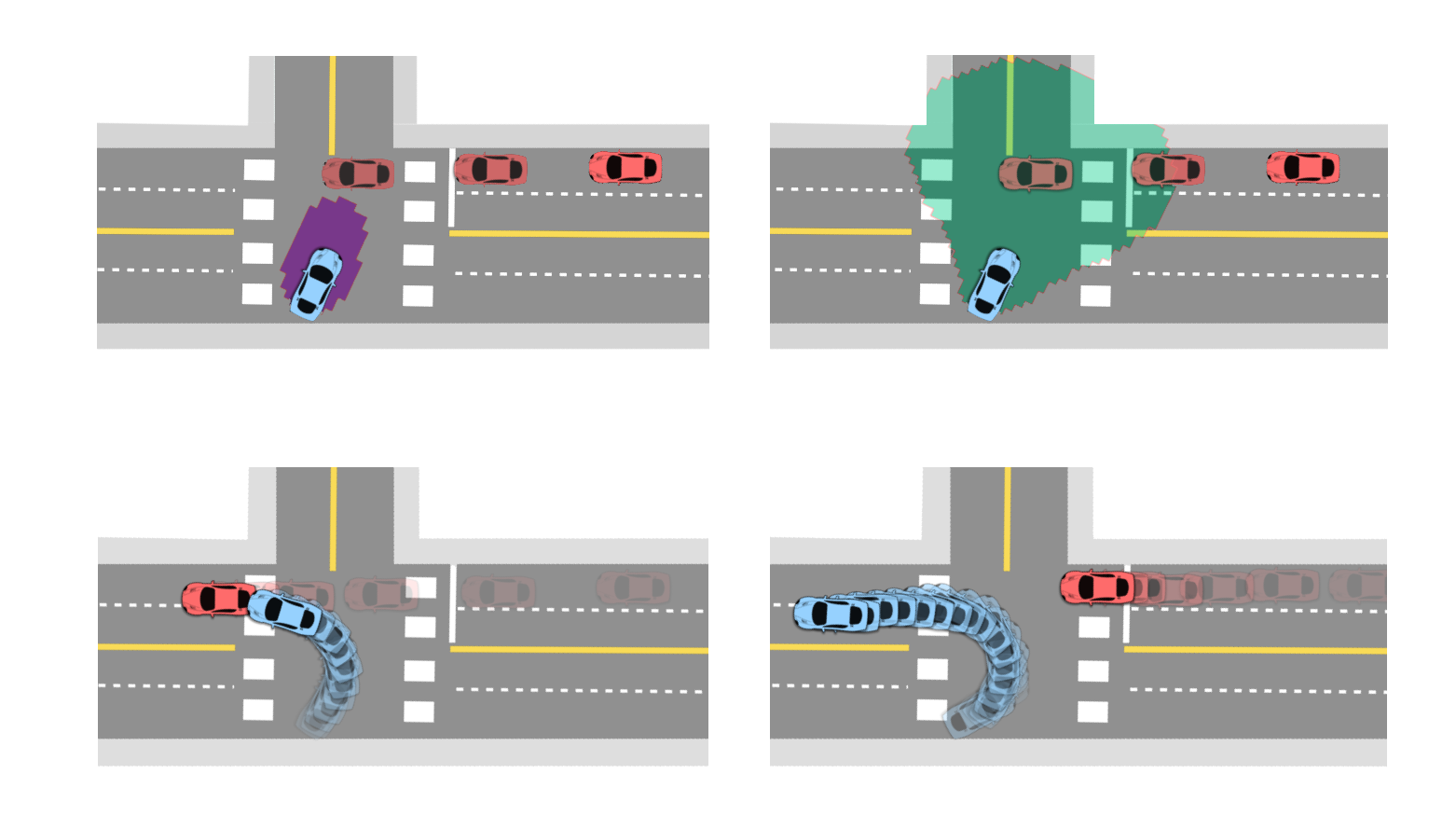}
\vspace{-1.5em}
\caption{
\textit{Top left:} The FRT, without using confidence estimation (purple), fails to alert the ego (red) vehicle that a collision is possible. \textit{Bottom left:} The ego vehicle learns the human's true intent (i.e., making a U-turn) too late and crashes into the human (blue) car. \textit{Top right:} The FRT of the human vehicle with confidence estimation (cyan).
\textit{Bottom right:} The use of confidence estimation allows the ego vehicle to detect a potential collision 2 seconds earlier than without confidence estimation, and safely stop before the stop line. This allows the human to complete the maneuver without any safety violations.}
\vspace{-2em}
\label{fig:uturn}
\end{figure}
When a robot operates in close proximity to humans, it often employs a model of the human's behavior to predict her future actions, and subsequently, her future states. Given the possible future states of the human, the robot may leverage online motion planning methods to plan around these moving obstacles, and generate real-time dynamically feasible and safe trajectories. However, when the predictions deviate from the true human behavior, the robot might confidently plan unsafe maneuvers and violate critical safety constraints.
%

To overcome this challenge, recent works have proposed estimating a confidence parameter in the human model based on how well the predictions align with the observed human behavior \cite{dfkConfidence}. 
The confidence parameter is then used to dilate the possible future actions of the human, essentially predicting every possible human action when the confidence is low. Even though promising, this approach is hard to scale for many human-robot applications because an update in the confidence parameter requires recomputing the future human states \textit{online}, which can be computationally demanding even for common, nonlinear dynamical systems.

In this work, we cast the computation of future human states as a Hamilton-Jacobi (HJ) reachability problem \cite{MitchellTimeDependent, lygeros2004reachability, bansal2017hamilton} and leverage recent advances in high-dimensional reachability analysis to quickly update human state predictions online.
Specifically, in reachability analysis, the future human states can be obtained by computing the \textit{forward reachable tube (FRT)} of the human -- the set of all states that the human can reach starting from its current state under predicted control actions.
Our key idea to update the FRT online is to compute a parameter-conditioned FRT \cite{borquez2022parameter} of the human, wherein a family of the FRT is computed and a member of this family can be obtained by specifying the confidence parameter value. 
Thus, the FRT can be updated online with a simple query of the parameterized family, corresponding to the current confidence estimate.

The parameterized FRT can be obtained by adding the confidence parameter (and other environment factors that might change online) as ``virtual states'' to the human dynamics and computing the FRT for the augmented system using the standard reachability tools. 
This, however, results in a high-dimensional reachability problem, especially for high-capacity predictive models that leverage semantic environment factors for accurate predictions, which now become additional parameters in the model that are only known online.
To overcome this challenge, we leverage DeepReach \cite{bansal2020deepreach} -- a reachability toolbox that builds upon recent advances in neural partial differential equation (PDE) solvers to compute high-dimensional reachable sets. DeepReach along with a parameterized FRT allows us to ensure safe human-robot interaction despite erroneous predictions.
%

To summarize, the key contributions of this work are:
(1) incorporating confidence estimation in high-capacity human prediction models, e.g., models based on deep neural networks.
The proposed framework allows us to exploit the predictive power of these models to plan efficient robot trajectories, yet ensure safety when the predictions cannot be trusted;
(2) developing a Hamilton-Jacobi reachability framework to update the model confidence and the corresponding safety assurances online for safe human-robot interaction.

\section{Related Work}
\label{sec:literature}
\noindent \textbf{Human Modeling and Prediction.} It is a common viewpoint that humans are rational agents, that is, that humans act with \textit{intent}. A common model used in human-robot interaction domains is the Boltzmann model, which captures the notion that humans are exponentially more likely to choose actions that maximize some reward function \cite{dfkConfidence},\cite{huSharp}, \cite{bobuLess}. 
However, reward functions that incorrectly specify human intent can lead to overly confident incorrect predictions.
Furthermore, the reward functions used to model the human's goals often fail to capture semantic information that impacts human decision-making. 
%
%
Specifically, in contexts such as autonomous driving, semantic information like stop signs or crosswalks shape how humans make decisions. One way to leverage semantic information and make predictions in the continuous action space is to use a neural network-based human model. 
These models have enabled inference and planning around human arm motion \cite{amor2014interaction, koppula2013anticipating}, navigation \cite{Ma_2017_CVPR, rosmann2017online}, and autonomous driving \cite{schneemann2016context, salzmann2020trajectron++, multipath} (see \cite{rudenko2019human} for a survey).
However, data-driven approaches are in general subject to incorrect predictions in scenarios not captured in the training data.
In this work, our goal is to ensure safe human-robot interaction despite erroneous predictions.

\vspace{0.2em}
\noindent \textbf{Safe Motion Planning.} The notion of safety in the context of human-robot interaction is well studied \cite{sadigh2016information, driggs2018robust}. 
Works in \cite{huSharp, leung2020infusing} use backward reachability to find the set of unsafe states and utilize them within a model predictive control framework to plan efficient trajectories.
Although some of these works \cite{huSharp} track human model confidence, the safety-enforcing backward reachable set is typically computed for a fixed set of parameters.
\cite{dfkConfidence, li2021prediction, ranSafety, bansal2020hamilton} add flexibility by precomputing a small discrete bank of reachable sets that reflect different potential beliefs of the human model. 
The system switches between these reachable sets based on which one fits the robots estimate best at runtime. 
However in practice, the parameters that affect the model predictions (and subsequently the unsafe set) are not known a priori and must be observed/estimated online, such as semantic information in the environment and the model confidence parameter. 
Any precomputed bank of reachable sets will suffer from being overly conservative in such scenarios. 
In this work, we propose a method to update such reachable sets online in an effective fashion.
%

\section{Problem Setup}
\label{sec:setup}
We consider a robot operating in a human occupied space. We assume that the robot has full knowledge of the environment, and the robot and human states.

\subsection{Agent Dynamics}
We model each agent as a dynamical system, where we denote the robot and human states as $\mathbf{x_R} \in \mathbb{R}^n$ and $\mathbf{x_H} \in \mathbb{R}^m$ respectively.
Their individual dynamics and controls are as follows:
\vspace{-0.7em}
\begin{equation}
\dot{\mathbf{x}}_i =  f(\mathbf{x}_i, \mathbf{u}_i) \; \; i \in [\mathbf{R}, \mathbf{H}]
\end{equation}
%
We also let $\xi(\tau; \mathbf{x}_i, \mathbf{u}_i(\cdot), t)$ denote the agent state at time $\tau$ starting at the state $\mathbf{x}_i$ at time $t$ and applying control $\mathbf{u}_i(\cdot)$ over the time horizon $[t, \tau]$.

The robot is assumed to have some objective or task, such as reaching a goal state, that it needs to plan and execute a trajectory for.
While the robot performs its task, it is imperative for it to never incur any safety violations.
We denote by $\mathcal{C}$ the set of states the robot should avoid to ensure safety, e.g., because they imply physical collisions with the human.
In this work, we will compute $\mathcal{C}$ via evaluating a forward reachable tube of the human. 
%
\vspace{0.2em}

\noindent \textbf{Running example:} \textit{
We introduce a running example for
illustration throughout the paper.
We consider a scenario where an autonomous car is interacting with a human-driven vehicle at a traffic intersection.
We model both agents in this scenario as extended unicycles where $\mathbf{\dot{x}} = [\dot{x}, \dot{y}, \dot{\theta}, \dot{v}]^{\intercal} = [v\cos{\theta}, v\sin{\theta}, u_1, u_2]^{\intercal}$. 
The vehicle controls are given by steering rate and acceleration.
The unicycle model is widely used in the literature for modeling autonomous vehicles \cite{salzmann2020trajectron++}, \cite{multipath}. Given a collision radius of $R_{col} = 1.5$ m, we define $\mathcal{C}$ as the positions of the autonomous vehicle that are within a distance of $R_{col}$ of the human vehicle.
}

\subsection{Human Prediction Model}
In human-robot interaction scenarios, the robot typically maintains a model of human behavior in order to aid in the prediction of their future states. In this work, we are particularly interested in the settings where the human motion predictors might be high-capacity models that use semantic information about the environment as an input (e.g., the roadgraph and traffic light state in the context of autonomous driving), along with the human states (and possibly their history) to generate continuous distributions over human controls.
We assume that at each time step $t$, the robot has a prediction for each time step over the prediction horizon $[t, t+T]$ in terms of multivariate Gaussian distribution over human control actions: 
\begin{equation}
\label{pred}
\mathbf{u}_H^{t:t+T} \sim \mathcal{N}(\mu^{t:t+T}, \Sigma^{t:t+T})
\end{equation}
Here, $\mu^{t:t+T}$ and $\Sigma^{t:t+T}$ are the vectors and matrices of appropriate dimensions that represent the mean and covariance for the human control actions from time $t$ to $t+T$. 
Such prediction representations are common in the literature, especially when the model is data-driven (e.g., \cite{salzmann2020trajectron++} and \cite{multipath}).

However, despite the benefits provided by semantic information and the use of continuous action distributions, predicted controls can differ significantly from the true actions taken by the human, especially in the scenarios that are scarce or not captured well in the training data. 
A robot leveraging this predictor can confidently plan unsafe motions in such situations. 
In this work, our goal is to ensure a safe human-robot interaction despite such predictor failures.
To account for such failures, we maintain a model confidence parameter indicating our level of trust in the prediction model \cite{dfkConfidence},\cite{bobuLess}. 
We use this model confidence to scale the covariance matrix in \eqref{pred} which dilates the predicted controls when model confidence is low.
We discuss our approach in more detail in Sec. \ref{subsec:confidence_estimation}.
\vspace{0.2em}

\noindent \textbf{Running example:} \textit{A variety of learned neural network-based methods have been proposed to predict the future actions of a human-driven car \cite{salzmann2020trajectron++}, \cite{multipath}, \cite{plop}, \cite{hongRoad}. 
For our case studies, we use Trajectron++ \cite{salzmann2020trajectron++}, which takes the semantic road information (such as lane boundaries and roadgraph), the ego agent history, as well as that of the surrounding agents as inputs, and outputs a distribution over future human actions.
Thus, the predicted mean and covariance matrices in \eqref{pred} are functions of these inputs.
}

\subsection{Safe Planning} \label{subsec:safe_planning}
Our overall goal is to plan efficient trajectories for the robot to complete its task while avoiding collisions with the human at all times based on an informed prediction of the human’s future motion. 
We leverage the idea of the forward reachable tube (FRT) to investigate how the set of occupancy states of the human evolves over time. 
Mathematically, the FRT of the human is defined as:
\vspace{-0.5em}
\begin{equation}
\label{eqn:frt}
\mathcal{V}(t) = \{y: \exists \mathbf{u}_H(\cdot), \exists \tau \in [t, t+T], \xi(\tau; \mathbf{x}_H, \mathbf{u}_H(\cdot), t) = y\}.
\vspace{-0.5em}
\end{equation}
Intuitively, the FRT represents the set of all states a human can reach under some control starting from their current state $\mathbf{x}_H$.
We limit conservative decision-making by considering a restricted set of physically feasible controls during the FRT computation, given by the human motion predictor in \eqref{pred}. 
These predictions are tempered by a model confidence parameter that expands the possible range of human control inputs when the confidence in the model's accuracy is low, and leave the original predictions in tact when model confidence is high.
To tractably compute the FRT, we also limit the human control actions to a bounded set of inputs by defining a probability threshold $\gamma$ on the distribution in \eqref{pred}.
We discuss this further in Sec. \ref{subsec:frt_computation}.
\vspace{0.2em}

\noindent \textbf{Running example:} \textit{
For our running example, the FRT consists of all possible states of the human-driven car under the Trajectron++ predictions.
Once, the FRT is computed, the potential future positions of the human can be obtained as: $\mathcal{K} = \{(x, y) \; | \; \exists v, \exists \theta, \; \mathbf{x}_H = (x, y, v, \theta) \in \mathcal{V}(t) \}$.
In other words, $\mathcal{K}$ is the projection of $\mathcal{V}(t)$ on the position states of the human.
Given the potential future states of the human, the unsafe states for the robot at time $t$ are given as: $\mathcal{C} = \mathcal{K} \bigoplus R_{col}$, where $\bigoplus$ denotes the Minkowski sum. 
$\mathcal{C}$ is essentially an expansion of the set $\mathcal{K}$ by $R_{col}$.
}

Since we will maintain a confidence in the human model, as this confidence evolve in time, the control bounds that parameterize the FRT also change. 
Consequently, the set of unsafe states, $\mathcal{C}$, that the robot needs to use during its planning change. 
Traditional methods for computing an FRT would require a re-computation of the entire FRT, which is generally a process too slow to do in real-time \cite{dfkConfidence}. 
On the other hand, using the worst case predictions (e.g., by considering all physically possible human actions) 
to generate FRTs can be overly conservative.
This necessitates the need to be able to update the FRT in real-time as predictions change.
We will do so by computing parameter-conditioned FRTs \cite{borquez2022parameter}.

\section{Overview: Hamilton-Jacobi (HJ) Reachability} \label{sec:reachability}
Our method for computing FRTs relies on Hamilton-Jacobi (HJ) reachability analysis.
In this section, we provide a quick overview of HJ reachability.

HJ reachability is a game-theoretic method for guaranteeing performance and safety for general nonlinear dynamics in the face of bounded disturbances and controls. 
In the context of this work, HJ reachability is used to compute the forward reachable tube, or the set of states that the human can reach given some set of starting states.
To compute the FRT for our system of interest, one must define a set of initial states $\mathcal{L}$ of the human\footnote{When $\mathcal{L}$ consists of a singleton state $\mathbf{x}$, we define a small neighborhood around $\mathcal{L}$ to be the set of initial states for numerical purposes.}. 
Next, we define a function $l(\mathbf{x})$ whose sub-zero level set yields $\mathcal{L}$, i.e., $\mathcal{L} = \{\mathbf{x}: l(\mathbf{x}) \leq 0 \}$.
The FRT is then computed by solving the following Hamilton-Jacobi-Bellman variational inequality (HJB-VI) \eqref{hjb}

\small
\vspace{-1em}
\begin{equation}
\begin{aligned}
\min \bigl\{ &\frac{\partial V(\mathbf{x},\tau)}{\partial \tau} + H(\mathbf{x},\tau, \nabla V(\mathbf{x},\tau)),~l(\mathbf{x}) - V(\mathbf{x},\tau) \bigr \} = 0 \\ &V(\mathbf{x},t) = l(\mathbf{x}), \; \tau \in [t, t+T]
\end{aligned}
\label{hjb}
\end{equation}
\normalsize
Here, $\nabla V(x, \tau)$ is the spatial derivative of the value function at time $\tau$ and  $H(x,\tau, \nabla V(x,\tau))$ is the Hamiltonian:

\small
\vspace{-0.5em}
\begin{equation}
\label{hamiltonian}
    H(x,\tau, \nabla V(\mathbf{x},\tau)) = \max_{\mathbf{u}} \nabla V(\mathbf{x},\tau) \cdot f_H(\mathbf{x},\mathbf{u})
\end{equation}
\normalsize
The solution to the HJB-VI in equation \eqref{hjb} is a value function $V(x, \tau)$, whose zero sub-level set at time $(t+T)$ describes the set of all states that the human could have potentially occupied from $t$ to $(t+T)$. 
This set of potential occupancies $\mathcal{V}(t)$ is formally written as
\begin{equation}
\mathcal{V}(t) = \bigl\{\mathbf{x} \; | \; V(\mathbf{x},t+T) < 0 \bigr \}
\end{equation}

Traditionally, these HJB-VIs are solved by discretizing the state space into a grid  and updating the value function at those grid points at each time step over the time horizon $[t, t+T]$ \cite{LST}. 
However, this discretization means that the computational time scales exponentially in dimensions, making it particularly challenging to update the FRT in real-time even for moderately dimensional systems.

A different approach is to solve the HJB equation via learning-based methods. DeepReach is a recently developed neural-network-based HJ reachability solver \cite{bansal2020deepreach}. 
DeepReach uses the HJB-VI itself to self-supervise the learning process of the solution.  
Advantageously, DeepReach does not discretize the state space and the solution time is more dependent on the problem complexity rather than the problem dimension. 
This enables the computation of highly parameterized reachable sets which can be used to update the reachable set online with a minimal amount of computation.

\section{Approach}

\subsection{Confidence estimation and update.} \label{subsec:confidence_estimation}
We assume that at each time step, the robot is able to observe the true action taken by the human $\bar{u}_H^{t-1}$. 
This means that, in hindsight, the robot can evaluate the likelihood of the true human action under its prediction model.
If the human took a low likelihood action, this implies that the predictive model failed to capture some aspects of the human's true intent. Drawing from the literature on model confidence estimations, we define a parameter $\beta \in \mathbb{R}$ to describe the level of confidence in the predictive model of the human. 

This parameter originates from the cognitive science discipline \cite{luce_1959}, where it captures the rationality of a human being modeled as a reward-maximizing agent.
High model confidence leads to the agent being exponentially more likely to take high reward actions and low model confidence transforms the agent's predicted distribution over controls to be uniform \cite{dfkConfidence}.

Given the continuous action space, we modify the definition of $\beta$. In this work, we define $\beta \in [\beta_{low}, \beta_{high}]$ which scales the covariance matrices $\Sigma^{t:t+H}$ in \eqref{pred}. 
$\beta_{high}$ is always set to be equal to 1, as having full confidence in the predictive model should capture the original predicted distribution over controls. 
$\beta_{low}$ is chosen to be a positive number less than $\beta_{high}$ based on the desired amount of spread in the human action that we want to safeguard against at low model confidence.
Incorporating, the model confidence in our predictions leads to:
%
\begin{equation}
\label{predUpdate}
u_H^{t:t+H} \sim \mathcal{N}(\mu^{t:t+H}, \frac{1}{\beta^t}\Sigma^{t:t+H}),
\end{equation}
where the covariance is scaled by $\frac{1}{\beta} \geq 1$. 

As opposed to previous works which only allowed a discrete set of $\beta$ \cite{ranSafety, dfkConfidence}, this work allows $\beta$ to take any value between $\beta_{low}$ and $\beta_{high}$. This has advantages as discrete values allow less expressiveness in the model confidence and can lead to large differences in the distribution over predicted actions, and subsequently the FRT, as $\beta$ changes.

To update $\beta$ online based on the observed human actions, we use a Bayes filter.
Specifically, the beliefs are initialized at $t=0$ such that $b_{-}^0(\beta_{low}) = b_{-}^0(\beta_{high}) = 0.5$. 
The update rule for $\beta$ is given by:
\vspace{-0.5em}
\begin{equation}
\label{beta}
b(\beta)_+^t = \frac{f(\bar{u}_H^t ; \mu^t, \frac{1}{\beta}\Sigma^t)b_{-}^t(\beta)}{\sum_{\tilde{\beta}}f(\bar{u}_H^t ; \mu^t, \frac{1}{\tilde{\beta}}\Sigma^t)b_{-}^t(\tilde{\beta})},
\vspace{-0.5em}
\end{equation}
where $b(\tilde{\beta})_{-}^t$ represents the a priori belief of the robot on the model confidence at time $t$, i.e., the probability of $\beta = \tilde{\beta}$ at time $t$.
$b(\tilde{\beta})_{+}^t$ represents the same probability a posteriori.
$f(\bar{u}_H^t; \mu^t, \frac{1}{\beta}\Sigma^t)$ refers to the Gaussian probability density function parameterized by mean $\mu^t$ and covariance $\frac{1}{\beta}\Sigma^t$ and evaluated at the observed action $\bar{u}_H^t$. Similarly to \cite{dfkConfidence}, we use the $\epsilon$-static transition model where at each time step, the belief is sampled from $b_0$ with some probability $\epsilon$ and otherwise retains the belief computed by \eqref{beta}. This can be expressed as
$
    b_{-}^t(\beta) = (1 - \epsilon) b_{+}^{t-1}(\beta) + \epsilon b^0_{-}(\beta)
$

Given a belief of both $\beta_{low}$ and $\beta_{high}$, the value chosen for $\beta$ to augment the predicted distribution is simply a linear interpolation with weights determined by the beliefs.
\vspace{-0.3em}
\begin{equation}
    \beta^t = \beta_{low}b_+^t(\beta_{low}) + \beta_{high}b_+^t(\beta_{high})
\end{equation}

\subsection{Collision Set Update} \label{subsec:frt_computation}
After the human predictions are modified by the model confidence parameter, the ego vehicle has knowledge of a Gaussian distribution over human controls, which can be used to compute the FRT and the robot unsafe states as discussed in Sec. \ref{subsec:safe_planning}.
However, most of the available methods to solve HJB-VI including DeepReach deal with a bounded range of possible controls.
We use a probability mass threshold $\gamma$ to trim the control distribution.
Specifically, for the $i$th entry, $u_i$, the range of control is given as [$\mu_i - \delta$, $\mu_i + \delta$], where $\delta$ is chosen such that the probability density function integrates to $\gamma$ over that range. 
\vspace{-0.5em}
\begin{equation}
    u_i \in [\mu_i - \delta, \mu_i + \delta] \text{  s.t. } \int_{\mu_i - \delta}^{\mu_i + \delta} f(u_i; \mu_i, \frac{1}{\beta}\Sigma_{ii})du_i = \gamma
    \vspace{-0.5em}
\end{equation}
%
Ultimately, this process provides the robot a deterministic range of human control actions $[u_{min}(\beta), u_{max}(\beta)]^{t:t+T}$ for each time step that varies as predictions or model confidence change.

We next address one of the core challenges of this work, i.e., updating the FRT online as the above control bounds change.
We treat $u_{min}(\beta)^{t:t+T}$ and $u_{max}(\beta)^{t:t+T}$ as additional virtual states for the human with zero dynamics and compute the FRT for the augmented system dynamics.
This results in a family of forward reachable sets characterized by the control bounds, which can then be used to query any set of control actions online and obtain the corresponding parameter-conditioned FRT \cite{borquez2022parameter}, and, subsequently, the update collision set $\mathcal{C}$.

Computing the FRT family requires solving a high-dimensional reachability problem, since now the human control bounds are also states in the system.
For this computation, we leverage DeepReach that is equipped to handle such problems.
As long as the confidence-modified predictions fall within the control range used during training, updating the collision set for the human driver can be done in real-time. 
\vspace{0.2em}

\noindent \textbf{Running Example:}
\textit{
The human FRT at any time depends on the current state of the human and the confidence-adjusted control bounds.
Without loss of generality, we can assume that the human starts at the origin and has zero initial heading.
To be able to compute the human FRT online for any values of the remaining parameters, we augment them to the human states.
This leads to new system dynamics:}
%
\begin{equation}
\mathbf{\dot{x}} =  \begin{bmatrix}\dot{\mathbf{x}}_H\\ \dot{v_{start}} \\\dot{\mathbf{u}}_{min}^{t} \\\dot{\mathbf{u}}_{min}^{t+T} \\ \dot{\mathbf{u}}_{max}^{t}\\ \dot{\mathbf{u}}_{max}^{t+T} \end{bmatrix} = \begin{bmatrix}f_H(\mathbf{x}, \mathbf{u}_H^\tau) \\ 0 \\\mathbf{0} \\ \mathbf{0} \\ \mathbf{0} \\ \mathbf{0} \end{bmatrix}, \quad \tau \in [t, t+T],
\end{equation}
\textit{with $\mathbf{x}$ being the 14D augmented state.
As such, we should add control bounds for every time step as a state; however, in practice we have found that a linear interpolation between the initial and final predicted controls serves as a good approximation.
Thus, $\mathbf{u}_H^{\tau} \in [\tilde{\mathbf{u}}_{min}^{\tau}, \tilde{\mathbf{u}}_{max}^{\tau}]$ approximately:
%
\begin{equation}
    \tilde{\mathbf{u}}_{min}^{\tau} = \left(1- \frac{\tau-t}{T}\right)\mathbf{u}_{min}^{t} + \left(\frac{\tau-t}{T}\right)\mathbf{u}_{min}^{t+T}
\end{equation}
$\tilde{\mathbf{u}}_{max}^{\tau}$ can be similarly defined.
}
\textit{Finally, the function $l(x)$ encoding the initial state of the robot is given by}
\begin{equation}
\begin{aligned}
\small{
l(\mathbf{x}) = \max\left(\| \begin{bmatrix}x \\y \end{bmatrix} - \epsilon_1\|_2,~|v - v_{start})| - \epsilon_2,~|\theta| - \epsilon_3\right),
    }
\end{aligned}
\end{equation}
\textit{
where $(x, y, v, \theta)$ represents the human-driven vehicle state as usual.
$\epsilon_1, \epsilon_2, \epsilon_3$ are small margins defined to have some mass in the initial set of states (for numerical purposes).
The FRT can now be computed using the HJB-VI as described in Sec. \ref{sec:reachability}.
Using DeepReach, we can solve the 14-dimensional computation that would not be possible using traditional reachability methods (typically limited to 6D). 
Once the parameter-conditioned FRT is learned, we can simply query the control bounds given by the current model predictions and the confidence parameter. Our simulation studies found that querying the learned parameter-conditioned FRT to check for collision took $0.0037 s$ on average, enabling real-time safety.
Correspondingly, the set $\mathcal{V}(t)$ describes the set of positions, orientations and velocities that the human can reach under confidence-adjusted control inputs, starting from $v_{start}$. 
The set of unsafe robot states can then be recovered as described in Sec. \ref{subsec:safe_planning}.
}


\section{Case Studies}

\begin{figure}[t]
\centering
\vspace{-1em}
\includegraphics[width=0.5\textwidth]{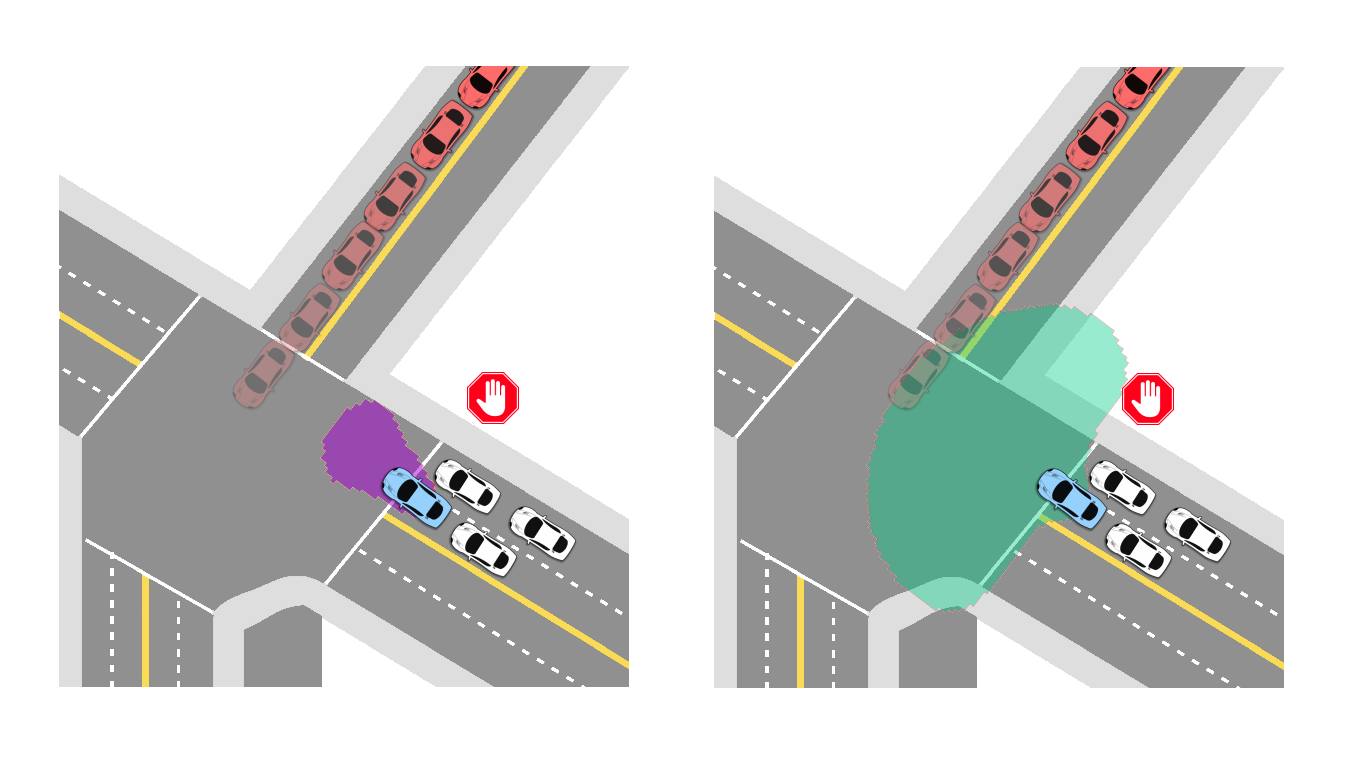}
\vspace{-2em}
\caption{\textit{Left:} Trajectron++ prediction of the human's (blue car) FRT. This causes the ego vehicle (red) to begin to slow down 5.8 m before the stop line. \textit{Right:} The confidence-based FRT causes the ego vehicle (red) to slow down 37.3 m before the stop line.}
\vspace{-2em}
\label{fig:stopline}
\end{figure}

To illustrate our approach, we use the NuScenes dataset \cite{nuscenes2019} to examine human-robot interaction for two traffic scenarios based on our running example. 
The NuScenes dataset has annotated semantic maps along with trajectory data for every 0.5 seconds, containing 1000 scenes. Among these scenarios, we identified edge cases (U-turn and running a stop sign) that are under represented in the dataset to find cases where predictive modules would have poor accuracy. 
In the first case study shown in Fig. \ref{fig:uturn}, the human-driven vehicle begins a U-turn maneuver from the right most lane as the autonomous vehicle approaches an intersection. 
In the second case study shown in Fig. \ref{fig:stopline}, the autonomous vehicle attempts to cross a protected intersection, but the human driver fails to yield and proceeds past the stop sign too quickly.

In these case studies, we use Trajectron++, a state-of-the-art human motion predictor, which utilizes the semantic information provided by NuScenes to better inform the distribution over future human actions. A publically available model pretrained on NuScenes data found in \cite{salzmann2020trajectron++} was used in the case studies in this paper.
The confidence in Trajectron++ at each timestep further parameterizes these distributions in the predictive model. 

For these case studies, the parameters were set such that $\beta_{low} = 0.03$ and $\gamma = 0.075$. The prediction horizon $T$ was 3 seconds, equal to the prediction horizon used for training the predictive module in \cite{salzmann2020trajectron++}. The values for $\gamma$ was chosen so that the resulting FRTs were similar to the FRTs computed in \cite{salzmann2020trajectron++} which were obtained via random sampling. $\beta_{low}$ was hand-tuned to achieve safe performance across the two example scenarios, without being overly conservative. 

Beyond our model of the agent dynamics, we also imposed hard constraints on the human control bounds to prevent unrealistic growth of the predictions. Namely, the magnitude of both the nominal and confidence-adjusted predictions were allowed to be 10 m/s$^2$ or 2 rad/s at most.

In both of these scenarios, we employ a simple planner for the ego vehicle -- the ego vehicle nominally moves in its lane at a constant speed.
If its current nominal trajectory intersects with $\mathcal{C}$ over the planning horizon (3 seconds), it decelerates to stop before the intersection if possible; otherwise, it decelerates maximally until it stops. 
Although the vehicle can use steering or more sophisticated planners for collision avoidance, we employ this simple planner to highlight the human-robot interaction aspects and because of the straight lanes in the chosen scenarios.



\subsection{U-turn} The first case study examined dealt with a human driver executing a U-turn at an intersection. 
Here, the ego vehicle is approaching the intersection at a speed of 21 m/s while
the human vehicle begins at rest in the rightmost lane and slowly begins a U-turn maneuver.

Figure \ref{fig:uturn} shows the difference in behavior when using confidence-based predictions versus trusting the raw Trajectron++ predictions. The belief evolution for the entire time horizon of the scenario is shown in Figure \ref{fig:uturnBel} with $\beta_{low}$ = 0.075. Here, as the human vehicle begins the U-turn from rest, Trajectron++ makes incorrect predictions the human will remain at rest/proceed slowly.
In actuality, the human vehicle turns quickly towards the left; consequently, the model confidence falls. 
However, as more of the maneuver is completed, Trajectron++ generates more accurate predictions and model confidence improves.

Notably, the confidence-based predictions enable the ego vehicle to detect a potential collision 52.46 m away from the stop line and stop safely with a deceleration of -4.2 m/s$^2$ over 4.9 seconds. In contrast, when the ego vehicle does not take into account model confidence, it fails to detect potential collision until it is 10.49 m away from the stop line. The ego vehicle tries to stop by applying the maximal deceleration of -10 m/$s^2$ for 2.1 seconds. Despite its best efforts, the ego vehicle ends up stopping in the middle of the intersection and colliding with the human.

\subsection{Running a Stop Sign}
In this next scene, the ego vehicle is crossing an intersection at a speed of 21 m/s with the right of way. Despite this, the human vehicle begins to enter the intersection early as the ego vehicle approaches.
Interestingly, this scene also has extra semantic information that leads to poor predictions -- the human car is surrounded by other cars that remain stopped behind the stop line. 
Trajectron++ uses the history of agents in close proximity to the human to inform predictions on their control actions and believing that the human car will also stop.
Thus, the robot immediately loses any confidence in the predictive model once the human begins to move forward. 
This is visualized in Figure \ref{fig:stopline} along with the FRTs with and without confidence. 
The belief is steadily regained as Trajectron++ determines the intent of the human. 

Using model confidence, the ego vehicle is able to detect a potential collision 47.8 meters away from the stop line, which is indicated by the white road line in Figure \ref{fig:stopline}. This allows the ego vehicle to gracefully stop before entering the intersection, decelerating at a rate of -4.61 m/$s^2$ for 4.55 seconds.  Without using confidence, the robot detects potential collision 5.83 m away from the stop line and crashes with the human despite maximal deceleration.

Each of the case studies illustrates that the updates of the FRT with respect to both changes in predictions and changes in model confidence result in a safer behavior. The FRTs in both scenarios, despite their differing appearance, were computed by the same value function. This shows that a single parameter-conditioned FRT can be used in different scenarios for safety updates.

\subsection{Effects of parameters $\beta_{low}$ and $\gamma$}

The choice of $\beta_{low}$ and $\gamma$ impacts the behavior of the car. Choosing these parameters can impact the degree of conservativeness of the vehicle. This section qualitatively explores the effects of different choices of these parameters.

\begin{figure}[ht]
     \centering
     \vspace{-2em}
         \includegraphics[width=0.5\textwidth]{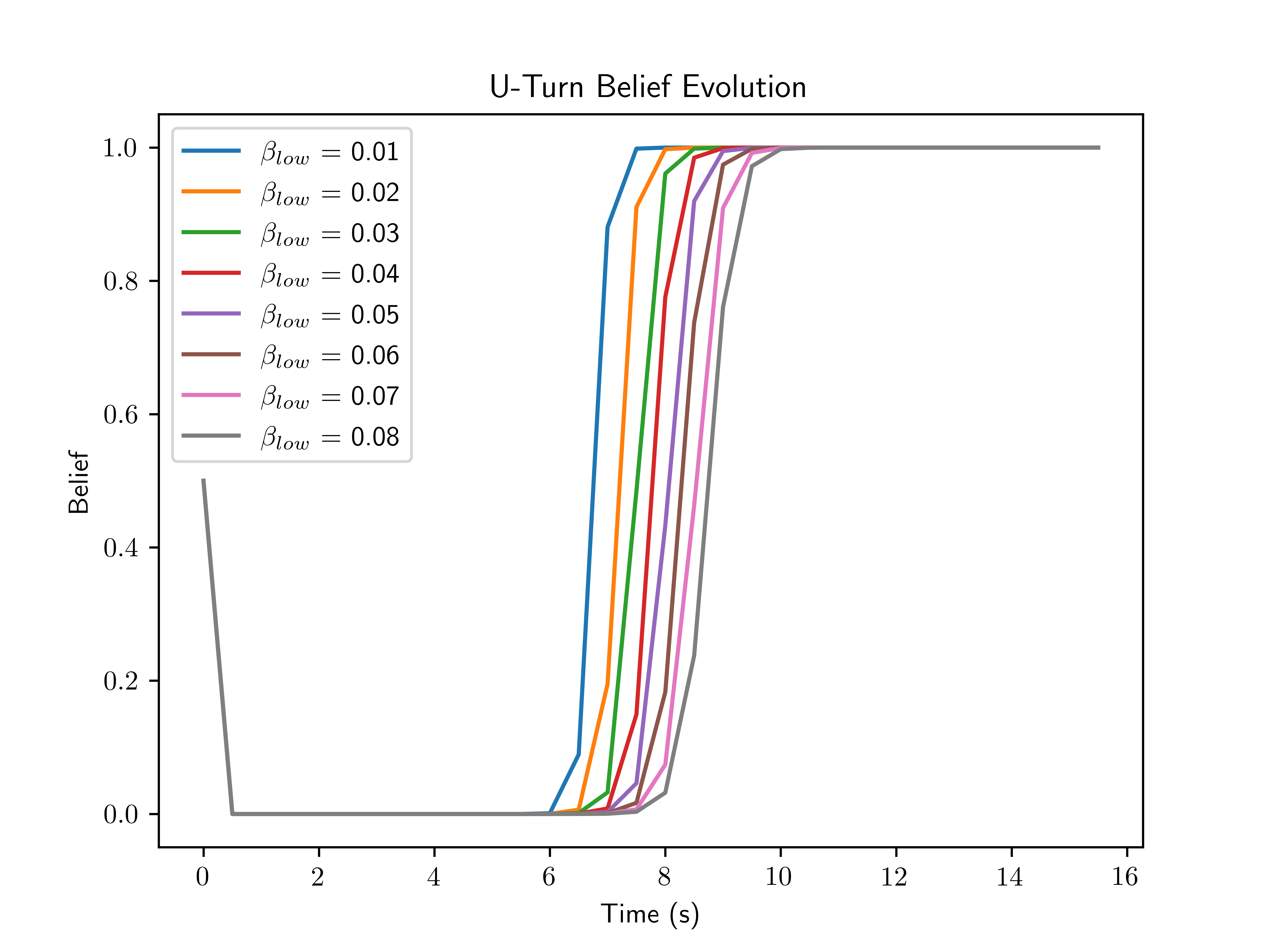}
         \vspace{-2em}
         \caption{As the car begins the U-turn maneuver, the confidence in the predictive model falls drastically. As the U-turn maneuver continues and the model starts to correctly predict the human actions, model confidence is restored. Different $\beta_{low}$ led to the confidence restoring at different points in time, with higher $\beta_{low}$ taking longer to regain confidence.}
         \label{fig:uturnBel}
\end{figure}

\textbf{Effect of $\mathbf{\beta_{low}}$:} Figure \ref{fig:uturnBel} shows that across the range of $\beta_{low}$ tested, the initial sharp decrease in belief remained identical, leading to the same qualitative behavior by the ego vehicle. However, the time at which the confidence was regained in the true model varied, with higher values of $\beta_{low}$ leading to confidence being restored later. This means the $\beta_{low}$ parameter, in addition to bounding the worst-case scaling of predicted actions, can be interpreted as encoding the rate at which the ego vehicle regains trust in it’s predictive model. We observe similar behavior for the stop sign scenario, so we omit the plot for brevity purposes.

\begin{figure}[ht]
\centering
\includegraphics[width=0.5\textwidth]{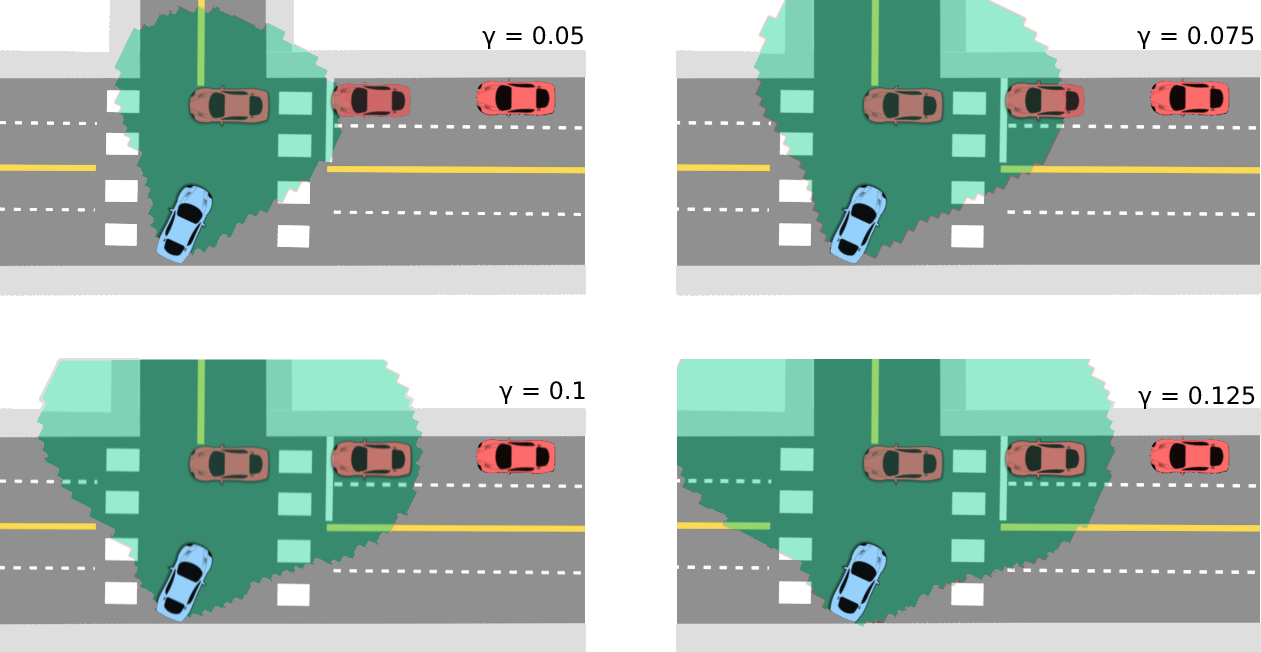}
\caption{Comparison of the parameter-conditioned FRT with different values of $\gamma$. The mean and covariance of the prediction is the same across all instances of $\gamma$.}
\vspace{-0.3cm}
\label{fig:gammaComp}
\end{figure}

\textbf{Effect of $\mathbf{\gamma}$:} The second parameter $\gamma$ is used to truncate the range of predicted human control actions. This means that the choice of $\gamma$ dictates how the predictive module influences the conservativeness of the ego agent. Figure \ref{fig:gammaComp} shows the effects of $\gamma \in \{$0.05, 0.075 0.1, 0.125$\}$.

Thus, the choice of $\gamma$ can serve a method of tuning the conservativeness of the ego vehicle. However, due to the open-loop nature of FRTs, high values of $\gamma$ may lead to overly conservative behavior. However, even with small values of $\gamma$, the inclusion of the confidence parameter helps the ego vehicle in making safer planning decisions.

\section{Discussion and Future Work}
\label{sec:conclusions}
In this paper, we propose to use parameter-conditioned reachable sets for the purpose of safe human-robot interaction. 
By making a time-varying parameterization of the human controls and model confidence, we demonstrated a real-time safety assurance updates for high-capacity human-robot interaction systems. 
Our two running examples grounded in the autonomous driving domain illustrated how our method's ability to quickly update the forward reachable tube of the human enables a safer human-robot interaction. 

In future, it would be interesting to apply our framework in the context of other human-robot interaction tasks, such as assistive manipulation and autonomous navigation. 
We will also investigate closed-loop human-robot interactions and operating around multiple humans.

\addtolength{\textheight}{-2cm}   






\bibliographystyle{IEEEtran}
\bibliography{./Bib/reachability, ./Bib/safe_motion_planning, ./Bib/bansal_papers, ./Bib/human_prediction, ./Bib/biblio}

\end{document}